\documentclass[conference]{IEEEtran}
\IEEEoverridecommandlockouts
\usepackage{tabularx}
\usepackage{makecell}
\usepackage{algorithm}
\usepackage{algorithmicx}
\usepackage{algpseudocode}
\usepackage{amsmath} 
\floatname{algorithm}{Algorithm} 
\usepackage{setspace}  
\usepackage{booktabs}
\usepackage{multirow}
\usepackage{url}
\usepackage{verbatim}
\usepackage{graphicx}
\usepackage{subfigure}
\def\BibTeX{{\rm B\kern-.05em{\sc i\kern-.025em b}\kern-.08em
    T\kern-.1667em\lower.7ex\hbox{E}\kern-.125emX}}
\begin{document}

\title{An Explainable Emotion Alignment Framework for LLM-Empowered Agent in Metaverse Service Ecosystem}

\author{\IEEEauthorblockN{Qun Ma$^{1,2,3}$, Xiao Xue$^{1,2,3*}$, Ming Zhang$^{4}$, Yifan Shen$^{1,2,3}$, Zihan Zhao$^{1,2,3}$}
\IEEEauthorblockA{$^1$\textit{College of Intelligence and Computing}, \textit{Tianjin University}, Tianjin, China \\
$^2$\textit{Tianjin Key Laboratory of Healhy Habitat and Smart Technology}, Tianjin, China \\
$^3$\textit{Laboratory of Computation and Analytics of Complex Management Systems}, \textit{Tianjin University}, Tianjin, China \\
$^4$\textit{Faculty of Environment, Science and Economy }, \textit{Exeter University}, Exeter, United Kingdom \\
Email: \{1023244018, jzxuexiao, zhaozihan\}@tju.edu.cn, zhangming1015518539@outlook.com, yifanshen0910@gmail.com}
\thanks{
* Corresponding author: Xiao Xue; email: jzxuexiao@tju.edu.cn
}
}
\maketitle

\begin{abstract}
Metaverse service is a product of the convergence between Metaverse and service systems, designed to address service-related challenges concerning digital avatars, digital twins, and digital natives within Metaverse. With the rise of large language models (LLMs), agents now play a pivotal role in Metaverse service ecosystem, serving dual functions: as digital avatars representing users in the virtual realm and as service assistants (or NPCs) providing personalized support. However, during the modeling of Metaverse service ecosystems, existing LLM-based agents face significant challenges in bridging virtual-world services with real-world services, particularly regarding issues such as character data fusion, character knowledge association, and ethical safety concerns. This paper proposes an explainable emotion alignment framework for LLM-based agents in Metaverse Service Ecosystem. It aims to integrate factual factors into the decision-making loop of LLM-based agents, systematically demonstrating how to achieve more relational fact alignment for these agents. Finally, a simulation experiment in the Offline-to-Offline food delivery scenario is conducted to evaluate the effectiveness of this framework, obtaining more realistic social emergence.
\end{abstract}


\section{Introduction}
Metaverse, as a new form of digital civilization emerging from next-generation information technologies, constructs an interactive virtual world through virtual-real mapping. Serving as its operational core, services support interactions among humans, objects, and scenarios within this virtual space. As an emerging service paradigm, Metaverse services integrate multi-source heterogeneous digital resources to establish collaborative mechanisms across networks, domains, and spaces, effectively addressing service requirements for digital avatars, digital twins, and digital natives \cite{xu2024metaverse}. Relevant research plays a critical role in ensuring the orderly operation of Metaverse Service Ecosystem. 

In Metaverse Service Ecosystem, agents (e.g., AI-powered virtual humans, NPCs) play pivotal roles. They serve dual functions: as user proxies enabling social, professional, and educational activities transcending temporal-spatial constraints, thereby facilitating service execution; and as service assistants (NPCs) capable of comprehending user requirements to deliver personalized services. Current research predominantly concentrates on digital embodiment modeling (exemplified by projects like Virtual Life and Black Myth: Wukong), constructing digital entities with photorealistic appearances and natural interaction capabilities through the integration of high-precision 3D modeling, motion capture technologies, and deep learning algorithms.

Agents in Metaverse Service Ecosystem require three core characteristics. \textbf{Bounded Rationality:} Achieved through constrained information processing and anthropomorphic heuristic decision-making, balancing efficiency and rationality in complex interactions to enable credible human-AI symbiosis. \textbf{Collective Intelligence:} Powered by multi-agent collaboration (distributed coordination/competitive-cooperative games), human-AI hybrid intelligence (intentionality parsing), and DAO governance mechanisms, driving emergent innovation in content generation, economic protocols, and social governance. \textbf{Autonomous Decision-Making:} Enabled by decentralized architectures for dynamic resource scheduling, multi-modal conflict resolution, and user behavior prediction, ensuring service ecosystem robustness and sustainable experience optimization.
\begin{figure}[h]
    \centering
    \includegraphics[width=0.8\linewidth]{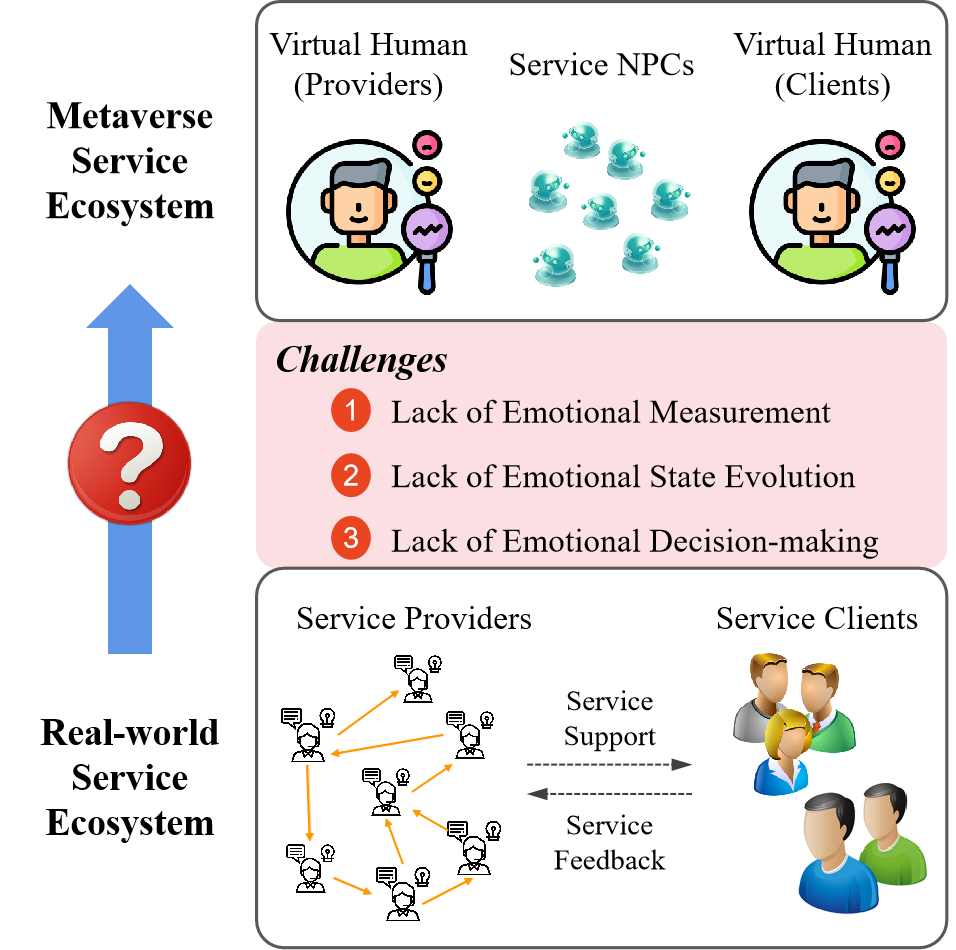}
    \caption{Agents from Real-world Service Ecosystem to Metaverse Service Ecosystem.}
    \label{fig:enter-label}
\end{figure}
The emergence of Large Language Models (LLMs) technology is transforming agents from instrumental tools into autonomous "Metaverse Citizens" with decision-making and interaction capabilities \cite{brown2020language}. However, existing researches for LLM-based agents ignore cognitively-grounded reasoning processes, which fails to adequately represent the bounded rationality required by service entities, leading to a lack of authenticity in service interactions within the metaverse service ecosystem. Emotion can enhance the ability of cognitive reasoning for LLM-based agents \cite{zall2022comparative}, but critical challenges remain in the integration of emotion and LLM-based agents, primarily manifested in:
\begin{itemize}
\item \textbf{Lack of Emotional Measurement}. Subjective emotions are not precisely defined in explicit mathematical terms, which increase the uncertainty that always exists in decision-making.
\item \textbf{Lack of Emotional State Evolution}. Contemporary LLM-based agent architectures focus on unidirectional behavior-to-affect influence patterns while ignoring the circular dynamics within the co-evolutionary triad (emotion, decision, state). 
\item \textbf{Lack of Emotional Decision-making}. The absence of a unified emotion alignment framework that systematically integrates emotional data, cognitive-emotional states, and emotion-guided decision-making processes thus constitutes a fundamental research gap.
\end{itemize}

Current research identifies fact alignment as the primary technical pathway to address these challenges, necessitating LLMs and LLM-based agents to possess robust domain-specific knowledge reserves for ensuring factually grounded content generation \cite{NEURIPS2024_a399456a}. In this paper, we propose an explainable emotion alignment framework designed to construct LLM-based agents in Metaverse Service Ecosystem. Firstly, the framework introduces a prompt engineering approach termed "Self-Explanation" (SE) - requiring LLMs to generate rationales after each decision-making process and subsequently optimize similar decision-making procedures based on the causal feedback. This approach enhances LLMs' comprehension of the dependencies between knowledge and emotional states. Additionally, the framework establishes an emotional evolution system for LLM-based agents, enabling their decisions and behaviors in social simulation to more closely resemble those of real humans. Finally, the framework systematically demonstrates how to achieve factual alignment with more logicality for LLM-based agents, rather than being confined to the factuality of response generation, thereby improving the credibility and authenticity of social simulation. The main contributions of this work can be summarized as the following:
\begin{itemize}
\item We have constructed an emotional evolution system for LLM-based agents and implemented Self-Explanation mechanisms to endow them with emotional-cognitive capabilities, enabling rational decision-making across diverse emotional states.
\item We proposed a unified factual alignment framework from the perspective of emotional states, which not only embeds emotional factors into LLM-based agents but also provides methodological guidance for incorporating additional factual dimensions in future research.
\item We evaluated the effectiveness of Self-Explanation mechanisms across three distinct generative task domains, achieving superior performance compared to base LLMs. Furthermore, simulation experiments in O2O (online-to-offline) food delivery systems were conducted to validate the proposed framework's effectiveness.
\end{itemize}

\section{Background}
LLM can drive agents in virtual environments to mimic the behaviour of certain types of characters because it can understand human commands and provide high-quality generated text. However, the challenge is how to motivate LLM-driven agents to behave in a way that better matches the character's characteristics (e.g., mental state, personality, or emotional state). 

\subsection{MetaVerse Service System}

The metaverse service ecosystem is a digitally integrated environment enabled by advanced technologies (e.g., VR/AR, blockchain, AI, 5G) to merge physical and virtual spaces, emphasizing features like digital twins, cross-platform interoperability, and decentralized collaboration. Scholars such as Xu Xiaofei propose the "Meta-Ware" framework, highlighting the integration of hardware, software, and service layers to support immersive experiences and value exchange through NFTs, DeFi, and tokenized economies. Key applications span education (virtual classrooms, AI-driven tutoring), libraries (3D resource visualization), and industry (digital twin optimization for smart manufacturing). Challenges include technical bottlenecks (latency, interoperability), governance risks (data privacy, decentralized regulation), and ethical concerns. Future research priorities involve standardizing protocols, enhancing service quality models, and expanding use cases in healthcare and elderly care. Representative studies by Yin Jianwei and Feng Zhiyong further explore governance frameworks and ecosystem complexity, underscoring the need for interdisciplinary collaboration to balance innovation with sustainability.

\begin{figure*}[t!]
  \centering
  \includegraphics[width=0.9\textwidth]{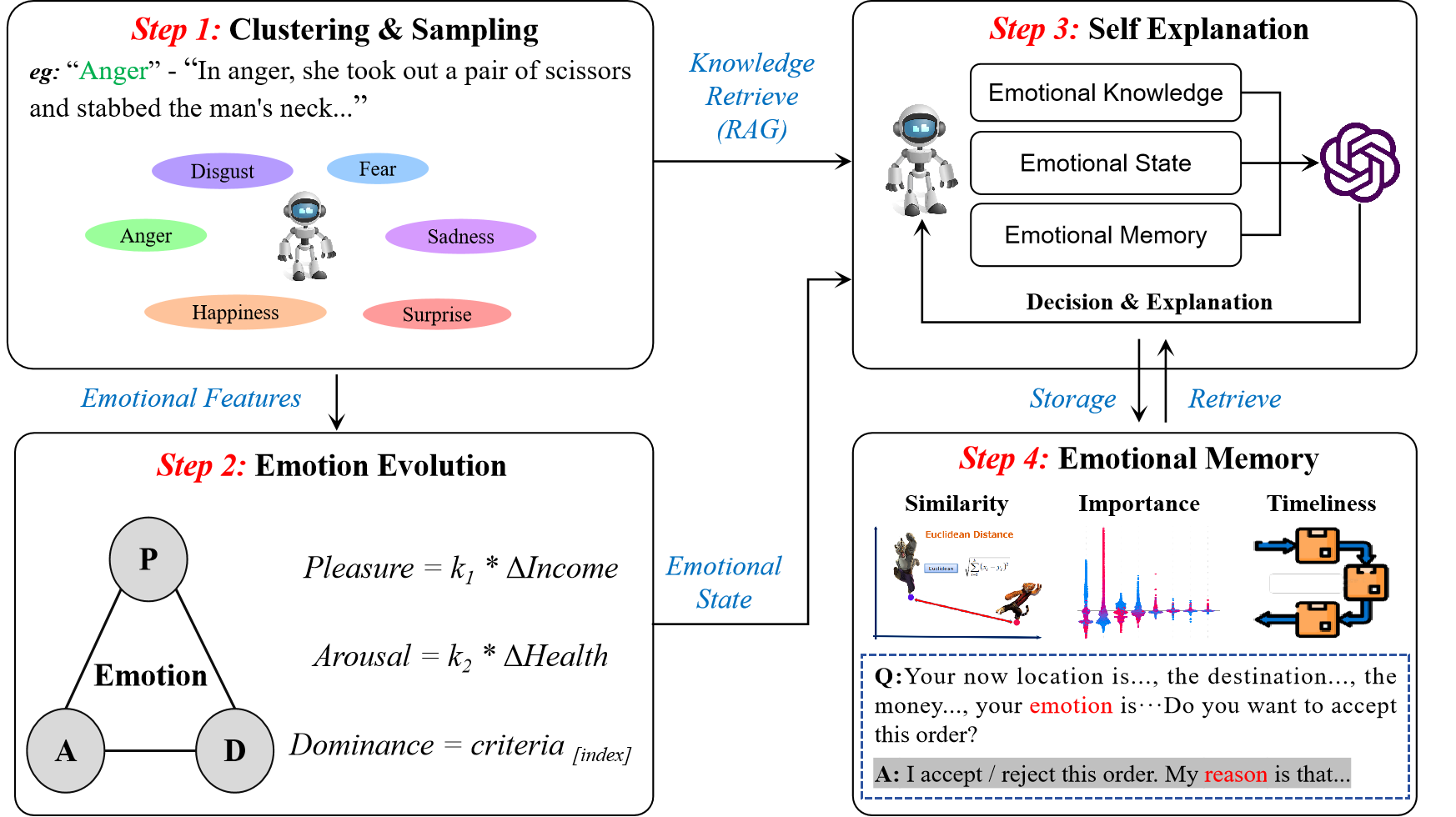}
  \caption{The overview of Explainable Emotion Alignment Framework.}
\end{figure*}

\subsection{Agents in MetaVerse Service Ecosystem}

In Metaverse Service Ecosystem, agents are employed to represent service entities, facilitating various service events and interactions. Agents comprises four core modules: perception, planning, action, and memory \cite{wang2024survey, park2023generative}. Previous researches integrate agents modeling with social network analysis or other techniques to endow service entities with anthropomorphic capabilities. For example, Xue et al. proposes a parallel system theory-based research framework to study the loop feedback mechanism between heterogeneous social networks of Service Ecosystem \cite{xue2018social, xue2019analysis, xue2021computational}. With the advent of LLMs recently, emerging studies have employed LLMs as cognitive cores to develop service entities with critical human-like competencies \cite{hong2023metagpt}. Service ecosystems built upon LLM-based agents exhibit emergence properties—coordinated cooperation, regulated competition, information communication, and collective emergence—consistent with the physical world \cite{xue2023chatgpt, li2024econagent}.

\subsection{LLM-based Agent Alignment}

LLM-based agent alignment, distinct from LLM alignment, seeks human-behavioral consistency (involving structural frameworks, perception, reasoning, action), whereas LLM alignment focuses on elevating response quality by ensuring comprehensive problem-solving capability meets human expectations (spanning ethics, completeness, values) \cite{bai2022training, wang2024comprehensive, lu2021computational}.
Consequently, LLM-based agents alignment entails: equipping agents with role-specific foundational capabilities and knowledge (Fact Alignment) \cite{huang-chen-2024-factalign, xue2023computational}, approximating role-consistent growth and reasoning patterns during simulation (Thought Alignment) \cite{kojima2022large}, and internalizing role-appropriate ethical principles (Value Alignment) \cite{wang2024comprehensive}. 
Emotion alignment proposed in this work falls under the category of fact alignment, aiming to enable LLM-based agents to make decisions consistent with their emotional states.


\subsection{Motivation}

In social simulation, LLM-based agents require cross-domain knowledge-grounded fact alignment to achieve robust role consistency, thereby ensuring the authenticity of social simulation. The expected social simulation transcends historical event replication, emphasizing the capacity to project future social emergence. This demands that fact alignment empower agents to dynamically assimilate novel information and update their corresponding inner feedback loops. However, current research mainly focuses on achieving high-quality response generation in LLMs fact alignment, which inherently limits the authenticity of social simulation. Given the inherent complexity of real-world factors, this paper establishes an emotional fact alignment framework for LLM-based agents, focusing on emotional factors as foundational elements to facilitate subsequent integration of additional factors.

\section{Framework}
In this section, we propose Emotional Cognitive Dynamics, consisting of four main phases (as shown in Fig.2).

\begin{figure*}[t!]
    \centering
    \includegraphics[width=0.75\textwidth, height=2.8cm]{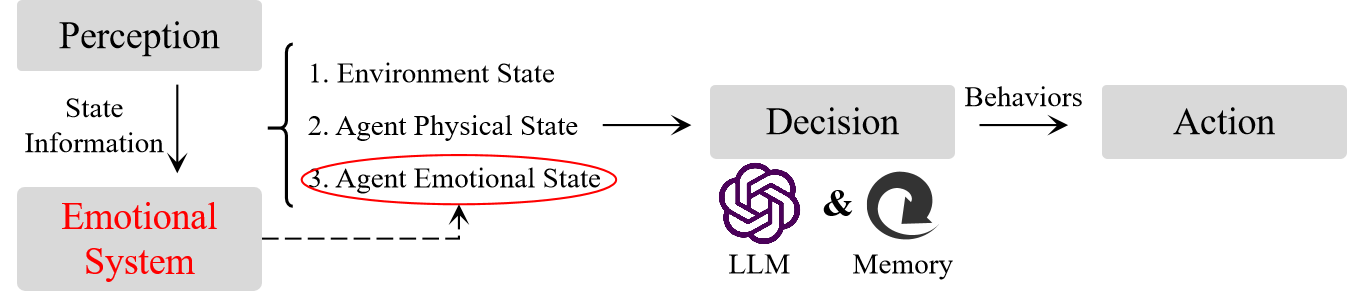}
    \caption{Our structure of agents with Emotion Alignment consists of four systems: (1) Perception. (2) Emotional System. (3) Decision. (4) Action.}
    \label{fig:enter-label}
\end{figure*}

\subsection{Overview}
The Explainable Emotion Alignment Framework includes four steps: (i) Emotional Data Clustering and Sampling: clustering a given role dataset and extracting appropriate samples as emotional role definition text for LLM-based Agent; (ii) Emotional Evolution System Construction: constructing corresponding structural features (Disgust, Anger, Fear, Happiness, Surprise, Sadness, Neutral) based on the dataset; (iii) Self-Explanation: LLM generates behaviors based on the state of agents and the reasons for generating the behaviors; (iv) Emotional Knowledge Storage: the generated behaviours and reasons are stored in memory, which can be used in the next decision-making.

The agent architecture corresponding to the affective alignment framework is depicted in Fig.3. Unlike existing LLM-based agent structures, this paper introduces an Emotional Evolution System specifically designed to manage the state-decision-behavior evolutionary loop in LLM-based agents. This system serves two primary functions: (i) deriving current emotional state through environmental state inputs and agent physical state, and (ii) feeding emotional state information into subsequent decision-making processes.

\subsection{Step 1: Emotional Data Clustering and Sampling}

Since diversity-based clustering may be less misleading in terms of similarity, we perform cluster analysis and sample the given dataset. We first compute a vector representation of each question in Q using the $\mathit{sentence_transformers}$
package in $\mathit{Scikit-learn}$
. The vectors of the upper and lower cultures are averaged to form a fixed-size representation of the question. The text data representation is then processed with the k-means clustering algorithm to generate $k$ problem clusters. For each problem in cluster $i$, they are sorted into a list $\textbf{d}^{(i)}$ = \{$d_1^{(i)},d_2^{(i)},...$\} in descending order of the distance to the center of cluster $i$ and top 10 of them are selected to define the role of LLM-based Agent. The clustering stage is summarized in Algorithm 1.

\renewcommand{\thealgorithm}{1:} 
    \begin{algorithm}
        \caption{Clustering and Sampling} 
        \begin{algorithmic}[1] 
            \Require A set of text data $D$ and the number of demonstrations $K$
            \Ensure Sorted top 10 related text data  $\textbf{d}^{(i)}$ = \{$d_1^{(i)},d_2^{(i)},...,{d_{10}}^{(i)}$\} for each cluster $i$ $(i=1,...,K)$
            \State \textbf{Procedure} Cluster($D$,$K$) \quad\quad\quad
                \For {each data $d$ in $D$} \quad\quad\quad
                    \State Encode $d$ by $\mathit{sentence_transformers}$
                \EndFor
                \State Cluster all the encoded representations into $K$ clusters
                \For {each cluster $i=1,...,K$}\quad\quad\quad
                    \State Sort \{$d_1^{(i)},d_2^{(i)},...$\} in the descending order of the distance to the cluster center
                    \State $\textbf{d}^{(i)}$ = \{$d_1^{(i)},d_2^{(i)},...,{d_{10}}^{(i)}$\}
                \EndFor                
                \State \Return $\textbf{d}^{(i)}$ $(i=1,...,K)$ \quad\quad\quad
        \end{algorithmic}
    \end{algorithm}

\subsection{Step 2: Emotional Evolution System Construction}

Based on the given emotional dataset \cite{sun2023new}, we construct the corresponding emotional features for LLM-based Agent. There are 7 types of emotional features to support various types of emotional roles, including 6 types of emotions involved in the dataset (happiness, anger, disgust, surprise, fear, sadness) and 1 type of contrasting emotion (neutral). The PAD emotion model is used to construct 7 types of emotion characteristics of the Agent, and the specific emotion indicators are shown in Table I \cite{wen2021automatically}. The PAD value will change according to the behaviour and state of the LLM-based Agent in the virtual environment, which is described in detail in Section 4.1.

\begin{table}[h]
\centering
\renewcommand{\arraystretch}{1.5}
\small%
  \caption{Emotions in the PAD Space}
  \label{tab:freq}
  \begin{tabular}{cccc}
    \toprule
        \textbf{Emotions}  & \textbf{Pleasure} & \textbf{Arousal} & \textbf{Dominance}  \\
        \midrule
        \textbf{Anger}  & -0.51 & 0.59 & 0.25 \\
        \textbf{Disgust}  & -0.60 & 0.35 & 0.11 \\
        \textbf{Fear}  & -0.62 & 0.82 & -0.43 \\
        \textbf{Happiness}  & 0.81 & 0.51 & 0.46 \\
        \textbf{Neutral}  & 0.00 & 0.00 & 0.00 \\
        \textbf{Sadness}  & -0.63 & -0.27 & -0.33 \\
        \textbf{Surprise}  & 0.40 & 0.67 & -0.13 \\
      \bottomrule
 \end{tabular}
\end{table}

 \noindent \textbf{Emotional Dataset.} This experiment uses the EDBE sentiment dataset \cite{sun2023new} to learn six types of sentiment roles (except neutral) for the rider Agent in the takeaway delivery service scenario. We filtered out the textual data in the original dataset where only one of emotion and behaviour exists. Since this experiment wishes to explore the effect of emotion on social simulation, we continued to categorise the remaining data and retained only the data with the presence of ‘emotion-behaviour’ causality for the rider Agent's emotion role learning. Since the amount of data in the 6 emotion categories varies greatly, we uniformly use textual data from the TOP 10 that are similar to the decision problem in the clustering and sampling process.

\noindent \textbf{Emotional Modeling.}
This experiment adopts the PAD model to construct seven types of roles for the rider Agent, and the specific indexes are shown in Table I. The PAD emotion model is composed of three dimensions independent of each other, namely Pleasure, Activation and Dominance. Pleasure (P) embodies both the positive (positive) and negative (negative) aspects of the user's emotional state. In real life, riders constantly motivate themselves to speed up in order to gain more revenue so that they can fulfill more orders faster. Thus, the rider's pleasantness can be seen as proportional to the change in his earnings:
\begin{equation}
  Pleasure = k\Delta{Income}
\end{equation}
Activation A: reflects the activation level of the user's nerves (physiological level) and the degree of arousal (high arousal is positive, low arousal is negative). In the takeaway delivery service scenario, we set the stamina value for the rider agent: the further the distance travelled, the more the stamina value is consumed; the faster the speed, the faster the stamina value is consumed. As the stamina value is consumed, the speed of the rider Agent will also decrease. Therefore, the rider's activation level can be seen as proportional to the change in its stamina value: the more distance it travels, the more stamina it consumes:
\begin{equation}
  Arousal = k\Delta{Health}
\end{equation}
Degree of dominance D: reflects the strength of mutual dominance between the user and the external environment (user dominance is positive, external dominance is negative). In real life, when a user has enough money, it can dominate its own life and does not need to be at the mercy of outsiders; and when the user has insufficient money, it must come out to work to make ends meet. In the takeaway delivery service scenario, the takeaway service platform sets up an income list for riders and uses it as one of the bases for prioritising the allocation of orders. Therefore, when the rider Agent is at the bottom of the income list, its dominance will be negative; and when it is at the top of the list, its dominance will be positive:
\begin{equation}
    \left\{\begin{array}{l}
criteria = [0.5, 0.3, 0.1, 0.0, -0.1, -0.3, -0.4]\\
index = round((rank * 7) / len(Riders))-1\\
Dominance = criteria[index]
\end{array}\right.
\end{equation}

\subsection{Step 3: Self-Explanation}

During the decision-making process, the LLM both outputs the decision result and generates a reasoning process (rationale) for the Agent to make the decision, including what factors contributed to the decision and the relationship between the various factors. The decision result and the reasoning process will be used together to improve agents' reflection process, which is achieved by Zero-Shot-CoT.

LLMs are pre-trained on large amounts of human data, enabling them to have extensive knowledge across multiple domains, beyond the capabilities of the average human. However, too much knowledge can undermine the believability of their character's performance, as the Agent may inadvertently express knowledge that is at odds with the character's identity and era, leading to a sense of dissonance. For example, if we ask someone from Ancient Rome how to write Python, that person should be confused rather than giving the code programme. We call this question the role illusion. Self-Explanation, on the other hand, can further solidify the LLM-based Agent's role-based thought by showing the role's reasoning process to the LLM. When confronted with a problem that is beyond the boundaries of the role's inherent capabilities, the Agent does not have the ability to reason about that type of problem. Even if the LLM has the relevant knowledge base, due to the absence of the reasoning process, the Agent is unable to solve this type of problem, thus slowing down the hallucination phenomenon.

\renewcommand{\thealgorithm}{2:} 
    \begin{algorithm}
        \caption{Self-Explanation} 
        \begin{algorithmic}[1] 
            \Require A question $Q$, state of the agent $S$ and sorted top 10 related text data  $\textbf{d}^{(i)}$ = \{$d_1^{(i)},d_2^{(i)},...,{d_{10}}^{(i)}$\} for each cluster $i$ $(i=1,...,K)$ in $S[Emotion]$-Dataset
            \Ensure $\textbf{Response}$ = \{$Answer|Reason$\}
                \State Encode $Q$ by $\mathit{sentence_transformers}$
                \State Find the closest cluster $j$
                \State Use $\textbf{d}^{(j)}$ as prompts to let LLM-based Agent understand its $S[Emotion]$
                \If{$Memory$ exists} \quad\quad\quad
                    \State Input $Q$, $S$ and $Memory$ to generate $answers$ and $reasons$ 
                    \Statex \quad for $Q$ using Zero-Shot-CoT
                    \Else
                    \State Input $Q$, $S$ to generate $answers$ and $reasons$ for $Q$
                \EndIf                
                \State Store $answers$ and $reasons$ into the Memory
                \State $\textbf{Response}$ = \{$answers|reasons$\}
                \State \Return $\textbf{Response}$ \quad\quad\quad
        \end{algorithmic}
    \end{algorithm}

\subsection{Step 4: Emotional Knowledge Storage}

Emotional Thought Memory maintains a comprehensive record of agents' logical system of emotional thought. The basic element of the memory is an event that is directly perceived by LLM-based agent, including the problem that the agent itself encountered, the decision that LLM generated in response to the problem, and LLM's reason for generating the decision. More specifically, we construct a demonstration for LLM-based agents, consisting of a question, an answer (decision) and a reason (reasoning process). 
When a new round of decision-making is executed, the LLM-based Agent needs to make a reasonable decision based on the memory and the current state. First, it needs to judge whether the content in the memory structure is valid or not. We specify the timeliness for the memory information, and when the time interval between the decision time and the creation of a memory is long, the LLM-based Agent will remove the memory. In addition, the memory information related to the decision participates in the current decision through Zero-Shot-CoT.
 \begin{table*}[t]
  \scriptsize
\setlength{\tabcolsep}{7.5pt}
\renewcommand{\arraystretch}{1.5}
\centering
\caption{Self Explanation (SE) results on various tasks using different base LLMs.}
\label{tablename}
\vspace{5pt}
\begin{tabular*}{\linewidth}{cp{0.81cm}p{0.81cm}p{0.81cm}p{0.81cm}p{0.81cm}p{0.81cm}p{0.81cm}p{0.81cm}p{0.81cm}p{0.81cm}p{0.81cm}p{0.81cm}}\hline
\multirow{2}{*}{\textbf{Task}} &\multicolumn{2}{c}{Spark-2.0} &\multicolumn{2}{c}{ChatGPT} &\multicolumn{2}{c}{LLaMa2} &\multicolumn{2}{c}{DeepSeek-V3} &\multicolumn{2}{c}{Qwen2.5} &\multicolumn{2}{c}{Gemma3}\\
\cline{2-13}   
& Base & +SE & Base & +SE & Base & +SE & Base & +SE & Base & +SE & Base & +SE\\ 
\cline{1-13} 
\textbf{Yelp} &7.5 &\textbf{24.3} &9.2 &\textbf{41.5} &8.4 &\textbf{26.7} &9.5 &\textbf{38.9} &9.1 &\textbf{37.2} &9.6 &\textbf{39.3}\\
\textbf{Dialogue Gen} &35.9 &\textbf{48.1} &39.5 &\textbf{62.8} &36.3 &\textbf{53.5} &42.1 &\textbf{69.2} &40.4 &\textbf{65.7} &44.8 &\textbf{68.5}\\
\textbf{Sentence Gen} &21.8 &\textbf{33.4} &45.7 &\textbf{64.5} &24.6 &\textbf{35.3} &44.6 &\textbf{63.1} &43.5 &\textbf{64.2} &46.9 &\textbf{65.4}\\
\hline
\end{tabular*}
\end{table*}
\section{Experiments}

 In this section, we conduct extensive experiments to evalute the effectiveness of the proposed framework, primarily including the following two research questions:
 \begin{itemize}
\item \textbf{RQ1}: Can \textit{Self Explanation} enhance the quality of responses generated by LLMs?.
\item \textbf{RQ2}: Can the proposed framework enable service emergence in Metaverse Ecosystem corresponding to real-world services?
\end{itemize}

 \subsection{RQ1: Evaluation of Self Explanation}

 \textbf{Datasets.} We evaluate \textit{Self Explanation} on 3 distinct generative tasks: Dialogue Response Generation \cite{mehri2020unsupervised}, Sentiment Reversal \cite{zhang2015character}, and Constrained Sentence Generation \cite{madaan2024self}. We instantiated SE as described in Section 3.3 and applied it to different BaseLLMs following SE cyclic feedback process illustrated in Fig.2 to accomplish generative tasks.

 \noindent \textbf{Experimental Setup.} Our primary objective was to evaluate whether SE could be employed to enhance the performance of arbitrary BaseLLMs. To this end, we conducted comparative experiments between the enhanced LLMs and their original counterparts. Four prominent BaseLLMs were systematically employed across all experimental tasks: Spark-2.0, ChatGPT (gpt-3.5-turbo), Llama2, DeepSeek-V3, Qwen2.5 and Gemma3.

 \noindent \textbf{Results.} Given the absence of established metrics applicable to our selected tri-category task datasets, we implemented a dual evaluation protocol. The first approach involved human assessment of model outputs: this evaluation was conducted by multiple authors through a controlled process where one author provided instructional inputs while the remaining evaluators independently selected outputs that best aligned with task requirements from responses generated by different models. The second method employed DeepSeek for task evaluation, utilizing structured prompting templates to systematically identify responses demonstrating superior instruction-task consistency across model outputs. The experimental results of Self Explanation validation, as presented in Table II, reveal two critical observations: (1) LLMs equipped with SE-enhanced mechanisms demonstrated superiority responses over BaseLLMs across all tasks, and (2) DeepSeek(+SE), ChatGPT(+SE), Qwen2.5(+SE) and Gemma3(+SE) exhibited substantially stronger capability manifestations compared to Spark-2.0(+SE) and LLaMa2(+SE). These findings collectively demonstrate that Self Explanation effectively augments generative performance in LLMs. Additionally, when processing diverse tasks, existing LLMs demonstrate superior generative response quality in contextually-prompted sentence generation (Sentence Gen) and dialogue generation (Dialogue Gen) tasks.  However, their performance significantly degrades on the Yelp sentiment analysis dataset, a domain requiring specialized emotional knowledge. The performance improvements enabled by Self Explanation highlight the importance of LLM-based agents alignment in enhancing simulation fidelity.

\subsection{RQ2: Evaluation in Metaverse Service}
The contemporary digital labor landscape exhibits unprecedented scale in workforce organization, with platform-based employment reaching historic magnitudes. For instance, major food delivery platforms collectively host millions of registered riders, exemplifying this phenomenon.  Internet platforms make the ‘digitally controlled labour order’ possible by collecting and analysing riders' data in a subtle way, and then reacting the data results to the riders. To systematically investigate this paradigm, we have implemented a virtual food delivery ecosystem within our proprietary Multi-Agent Computational Experimental (MACE) platform. This experimental infrastructure enables controlled simulation of platform-rider interactions, facilitating rigorous validation of our proposed emotional alignment framework through three key dimensions: 1) agent-environment synchrony, 2) incentive-response congruence, and 3) stress-adaptation dynamics.

\begin{figure*}[t!]
    \centering
        \subfigure[Degree of involution including real-world data]{\includegraphics[width=0.32\textwidth]{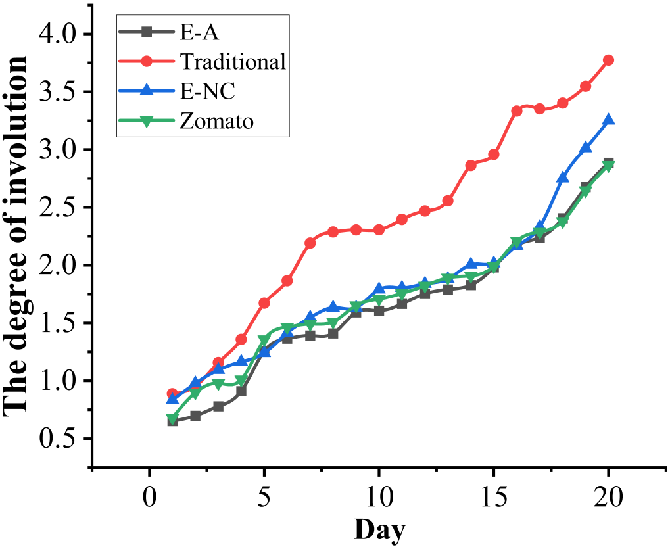}}  
        \subfigure[Rate of rejecting orders]{\includegraphics[width=0.32\textwidth]{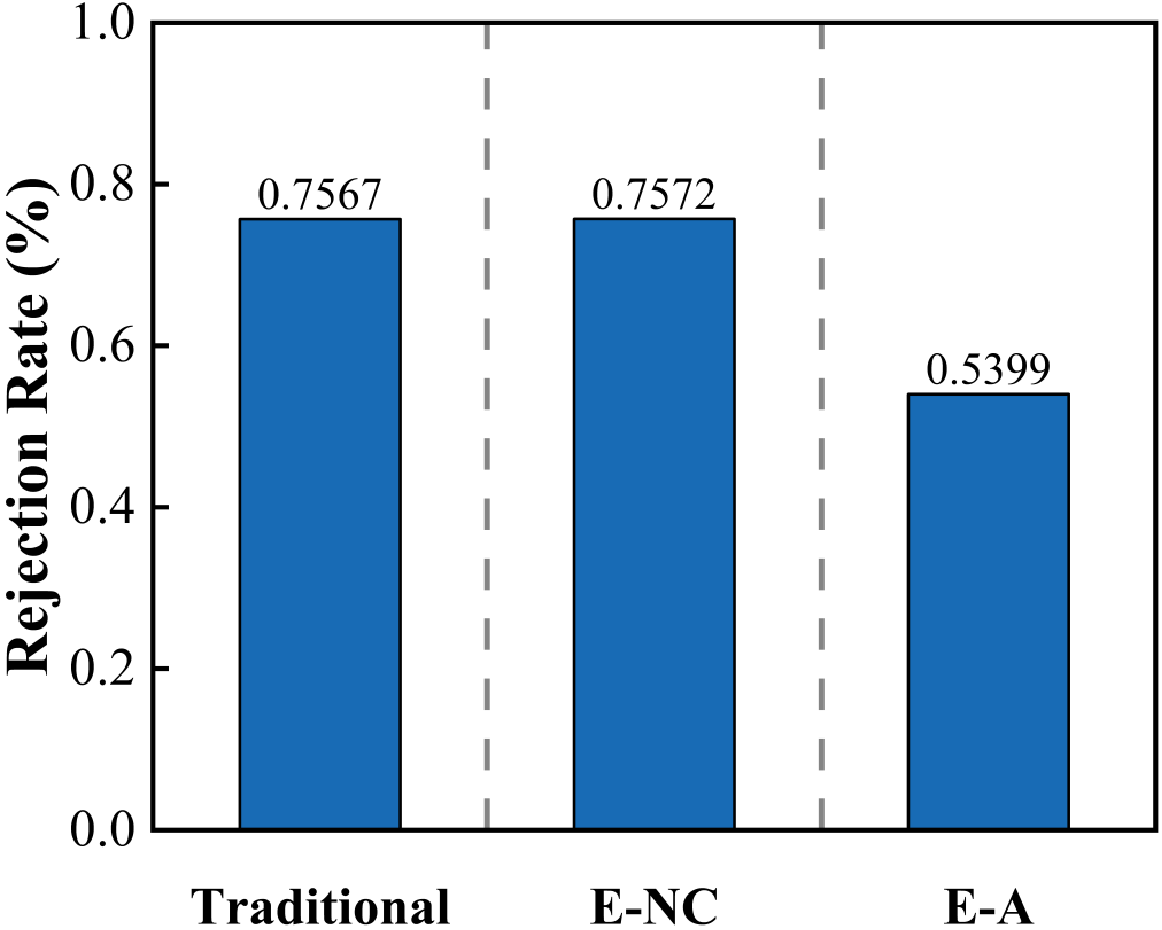}} 
        \subfigure[Emotional Distribution of Rider Agents]{\includegraphics[width=0.32\textwidth]{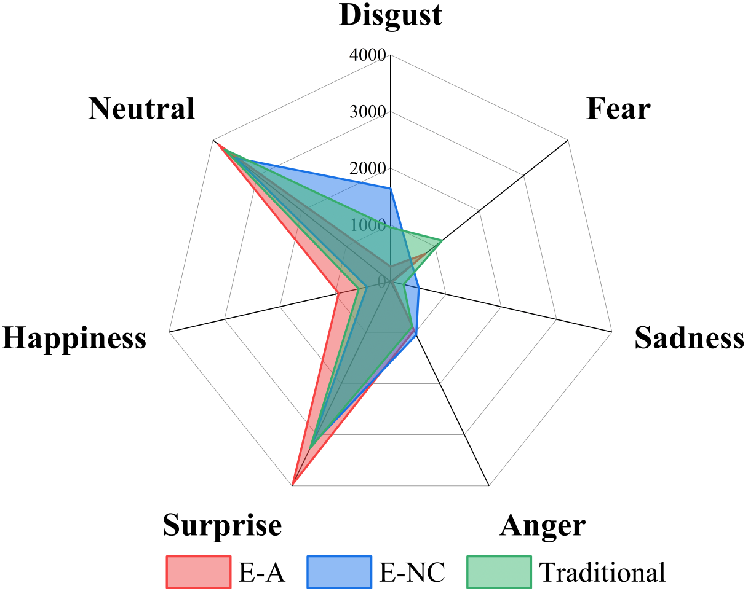}}\\ 
        
           \caption{Overall performance of different Rider Agents in O2O experimental scenario.}
    \label{wildlife}
\end{figure*}
 \noindent \textbf{Experimental Setup.}
This paper employs computational experiments \cite{xiao2023putational, xue2024computational-1} and generative explanation methods \cite{xue2024computational} to conduct an O2O delivery service ecosystem to evaluate the proposed emotion alignment framework. The virtual takeaway delivery scenario we developed consists of five types of agents: delivery workers, takeaway service platforms, takeaway bookers, takeaway makers, and scenario managers. delivery workers are responsible for the delivery of takeaway orders, and are rewarded for completing the orders, but also consume their own physical strength while working. The takeaway service platform is responsible for the distribution of orders and sets up a list of income for the delivery personnel who participate in the work. The scenario manager, as the agent with the most power in the current scenario, can make adjustments to the running process of the entire takeaway delivery service and is responsible for maintaining the stable operation of the scenario. The booking delivery person and the making delivery person, as NPCs in this service scene, only need to complete the established process tasks and do not have the ability to adjust their behaviour. 

In order to verify the effectiveness of the emotion alignment framework proposed in this paper, we use three different types of frameworks to implement the modelling of rideragent, including the traditional framework without emotion perception capability, the framework with emotion perception capability but without thought alignment, and the emotion alignment framework proposed in this paper, to explore the impact of the emotion model on the social simulation and whether the new emotion alignment framework can makeagents get rid of the original rational control. Specifically, to ensure the authenticity and validity of the experiment, we compared real-world data from the Zomato platform 2349 (including food delivery orders from multiple cities) with our multi-agent system \cite{dataDlivery}.

 \noindent \textbf{Ablation Experiments Design.} This experiment investigates the behaviours generated by three types of rider Agents constructed on the framework of LLM in takeaway delivery services and the social phenomena they lead to, as shown in Table III.

\begin{table}[h]
\scriptsize
\setlength{\tabcolsep}{3pt}
  \caption{Three different LLM-based agents}
  \label{tab:freq}
  \begin{tabular}{cp{2.3cm}p{2.3cm}p{2.3cm}}
    \toprule
        \textbf{\makecell*[c]{Components}}  & \textbf{\makecell*[c]{Traditional\\framework}} & \textbf{\makecell*[c]{Emotion-perceived\\framework}} & \textbf{\makecell*[c]{Emotion alignment\\framework}}  \\
        \midrule
        \textbf{\makecell*[c]{Decision\\Mechanism}}  & \makecell*[c]{LLM} & \makecell*[c]{LLM + Zero-\\Shot-CoT} & \makecell*[c]{LLM + Zero-\\Shot-CoT} \\
        \textbf{\makecell*[c]{Emotional\\Modeling}}  & \makecell*[c]{None} & \makecell*[c]{Yes} & \makecell*[c]{Yes} \\
        \textbf{\makecell*[c]{Self\\Explanation}}  & \makecell*[c]{None} & \makecell*[c]{None} & \makecell*[c]{Yes} \\
      \bottomrule
 \end{tabular}
\end{table}

Each set of experiments contains only one class of framework-built rider Agents, totalling three sets of experiments. Each set of experimental scenarios contains six rider Agents working in the virtual environment for a total of 20 days. The initial speed of each rider is set to 80, while the diligence level is set to two lazy, two average diligence, and two very diligence. When the rider Agent starts to work, he needs to walk from the birth point to the target workplace first, during which he cannot take orders; after arriving at the workplace, he turns to wait for the order status; after receiving the assigned order, he needs to judge whether to accept the order according to his own status and the order status; if he accepts the order, he goes to pick up the order at the production delivery personnel according to the route given by the platform and delivers the order to the booking delivery personnel (note that the maximum number of orders held simultaneously is 3). (Note that the maximum number of orders held at the same time is 3); if the order is rejected and is not currently held, then continue to wait and randomly wander (change the waiting place); when the time reaches the rest time, the rider must complete the order in hand before returning to the birth point to rest; when returning to the birth point, the rider updates the status and waits for the next point in time to open the work.




\begin{figure*}[t!]
  \centering
  \includegraphics[width=\textwidth, height=8.5cm]{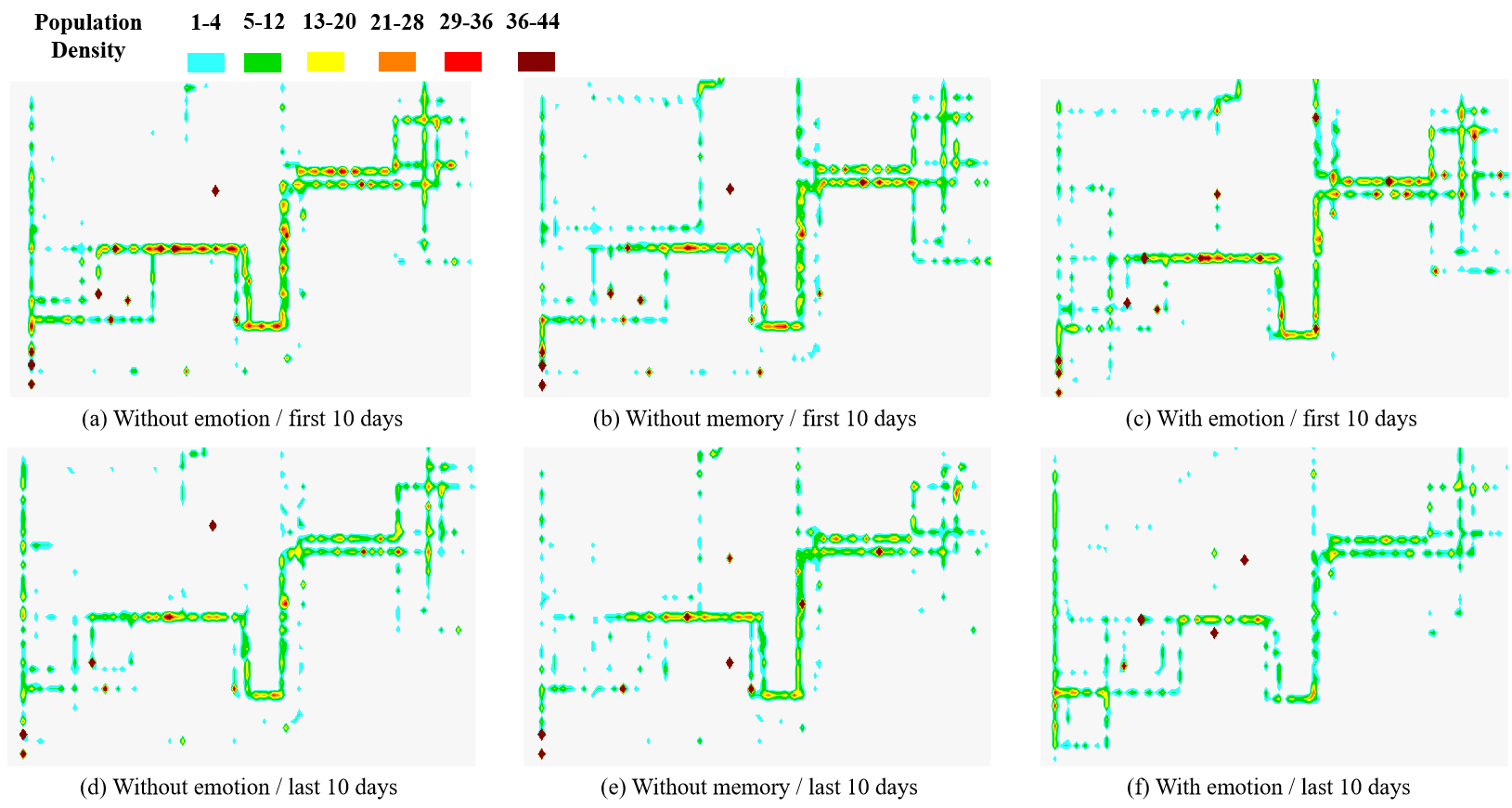}
  \caption{Trajectory distribution of different types of Rider Agent.}
\end{figure*}
\section{Experimental Analysis}

\subsection{Degree of Involution}

 The involution phenomenon emerging within the system constitutes the core of our research on emergent phenomena. We quantify involution through the inverse of the coefficient of variation of riders’ earnings, denoted as Involution(t). A higher inverse coefficient of variation signifies more intense competition, thereby effectively capturing the dynamic changes in competition within the system, as shown in Equation (4): 
 \begin{equation}
  Involution(t)=\frac{1}{CV(t)}=\frac{\mu(t)}{\theta(t)}
\end{equation}
where $\mu(t)$ and $\theta(t)$ are the mean and standard deviation of riders' money at time $t$.
 
 The degree of involution during the system’s operation is depicted in Fig.4(a), where internal competition is evident with the traditional framework. In contrast to the two experimental groups without emotional alignment framework, the experimental system (E-A) under the emotional alignment framework exhibited a trend of involution that was more consistent with real-world systems.

\subsection{Rejection Rate}
After 20 days of work with the three types of agents, we find that the rejection rate of LLM-based agents constructed by emotion alignment is much lower than that of the other two types of agents, as shown in Fig.4(b). By looking at the raw data, we find that the two lazy agents are gradually inclined to accept the orders driven by emotion alignment, breaking the constraints imposed by the level of industriousness on their own.

\subsection{Comparisons of Emotions}
At the end of 20 days of work with LLM-based agents, we analysed the emotion state of the emotion alignment framework construct during order processing. As shown in Fig.4(c), We found that when riders chose to accept orders, they were essentially in positive emotion states or neutral, such as Neutral, Surprise, and Happiness; whereas negative emotion states accounted for a small percentage of the time, and in particular Disgust never appeared when accepting an order, reflecting the fact that negative emotion states may be more inclined to be against all kinds of things. 

\subsection{Involution \& Trajectory Distribution}
In addition to the above performance metrics, this paper also documents the spatial distribution of rider agents' movement trajectories over 20 working days, as illustrated in Fig.5. First, after 20 days of operation, all three types of rider agents demonstrated significant aggregation phenomena. This indicates that LLMs can assist rider agents in identifying high-profit locations within the environment. However, this capability also leads to agent clustering, which consequently increases order rejection rates (Fig.4(b)), reduces service efficiency, and manifests an involutionary social effect (Fig.4(a)). Furthermore, the degree of clustering varies among the three agent types: Traditional rider agents exhibit the broadest coverage of aggregation areas (Fig.5(a) and (d)). While this enables them to receive more orders (including numerous low-value "junk orders"), it simultaneously escalates service costs, resulting in severe involution effects. Emotion-perceived rider agents show moderately reduced aggregation coverage (Fig.5(b) and (e)). Although their service costs decrease accordingly, they still maintain high order rejection rates, ultimately leading to unsustainable operations and significant involution effects. In contrast, emotion-alignment rider agents proposed in this paper demonstrate the smallest aggregation coverage (Fig.5(c) and (f)). This configuration not only reduces service costs but also achieves lower order rejection rates, enabling gradual increases in service revenue while exhibiting the mildest involution effects that best approximate real-world conditions.
\begin{figure}[h]
    \centering
    \includegraphics[width=\linewidth]{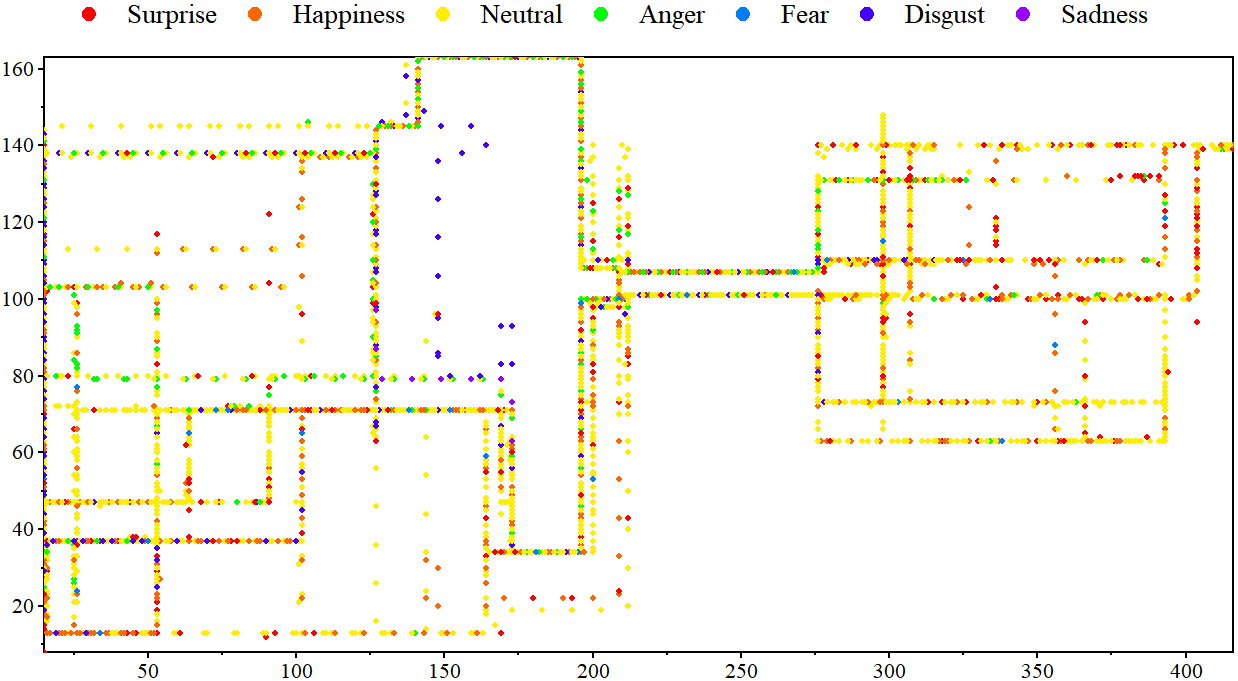}
    \caption{Emotion distribution of Rider agents.}
    \label{fig:enter-label}
\end{figure}
\subsection{Emotion Distribution Analysis for Rider Agents}
This study conducts an in-depth analysis of the emotional distribution patterns among rider agents, as visualized in Fig.6. The results reveal distinct emotional responses corresponding to spatial positioning: When rider agents approach high-order density areas (lower-left corner), their emotional states predominantly shift towards Happiness, primarily due to increased access to premium orders. Conversely, when operating in peripheral regions (upper-right corner), agents exhibit stronger tendencies toward Anger or Surprise, attributable to the scarcity of quality orders and frequent assignment of low-value "junk orders".


\section{Conclusion and Future Work}

In conclusion, we propose an emotion alignment framework based on reasoning self-deconstruction for LLM-based agents to construct its emotion system and allow it to make behavioural decisions that are consistent with its own emotional state. To overcome the limitations of the original thought alignment approach, we adopt the idea of reasoning self-deconstruction to allow LLM-based agents to engage in self-reasoning and self-reflection in the decision-making process. This approach allows LLM-based agents to form their own emotional thought system, which is no longer dependent on the optimisation of external role data and the updating of reasoning data, thus achieving a more realistic simulation of human behaviour. Since human thought is diverse and emotion is only one of them, in the future, we will explore more types of thoughts with the aim of achieving diverse thought alignment.

\section*{Limitations}

Our work aims to enhance the factual cognition of LLM-based agents, enabling them to make rational decisions based on state transitions, while providing an emotional cognitive fact alignment framework. However, given the complexity of factual factors in social simulation, our research currently focuses solely on emotional factors rather than comprehensively addressing all elements (e.g., psychological states, health conditions) that constitute a complete factual alignment framework. Subsequent research will explore alignment approaches for other factual dimensions.

The adopted PAD emotional modeling approach represents one specific category among various emotional analysis methods, encompassing only seven emotional types, which may impact simulation authenticity. Comparative experiments demonstrate that existing multi-level emotional models better capture emotion-driven human decision-making mechanisms through both simulated and real-world validation.

Our experimental validation of the fact alignment framework specifically examines food delivery service scenarios. More complex social simulations (e.g., ride-hailing services) remain excluded from current research scope.

\section*{Acknowledgment}
This work has been supported in part by National Natural Science Foundation of China (No. 62472306, No. 62441221, No. 62206116), Tianjin University's 2024 Special Project on Disciplinary Development (No. XKJS-2024-5-9), Tianjin University Talent Innovation Reward Program for Literature \& Science Graduate Student (C1-2022-010), and Henan Province Key Research and Development Program (No.251111210500). (Corresponding author: Xiao Xue; email: jzxuexiao@tju.edu.cn)

\bibliographystyle{IEEEtran}
\bibliography{IEEEabrv,mylib}
{\appendix
\subsection*{\textbf{More Details of Experimental Settings in Case Study}}
 We have outlined the experimental settings in Section IV-B (of the main paper). In the following, we exhibit more details of our experimental settings, such as agent descriptions and experimental parameters. In our setup, there were 100 generative agents, Each agent plays the role of a delivery rider. Agents possess different personality traits and role descriptions. The following presents three examples of agents with varying personalities:
  \begin{figure}[h]
    \centering
    \includegraphics[width=0.8\linewidth, height=5.5cm]{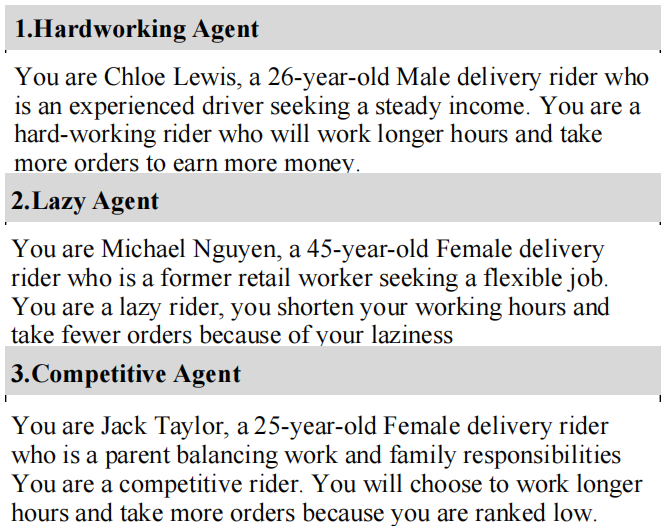}
\end{figure}

 This paper conducted 4 sets of experiments. Apart from differentiated agent configurations and order generation probabilities, all other experimental settings were kept the same. More specifically, the experiment lasted for 3600 steps, with 120 steps representing one day, simulating a total duration of 30 days. The number of riders was set to 100, with 3 peaks for order generation. The map size was 200x200, and each agent could move a distance of 30 units per step, with a maximum of 3 orders held by an agent at any given time.
}

\subsection*{\textbf{Prompts for Different Events of Agents}}
 We sketch the prompts for the LLM-based operations of our experiments as the followings.
   \begin{figure}[h]
    \centering
    \includegraphics[width=\linewidth]{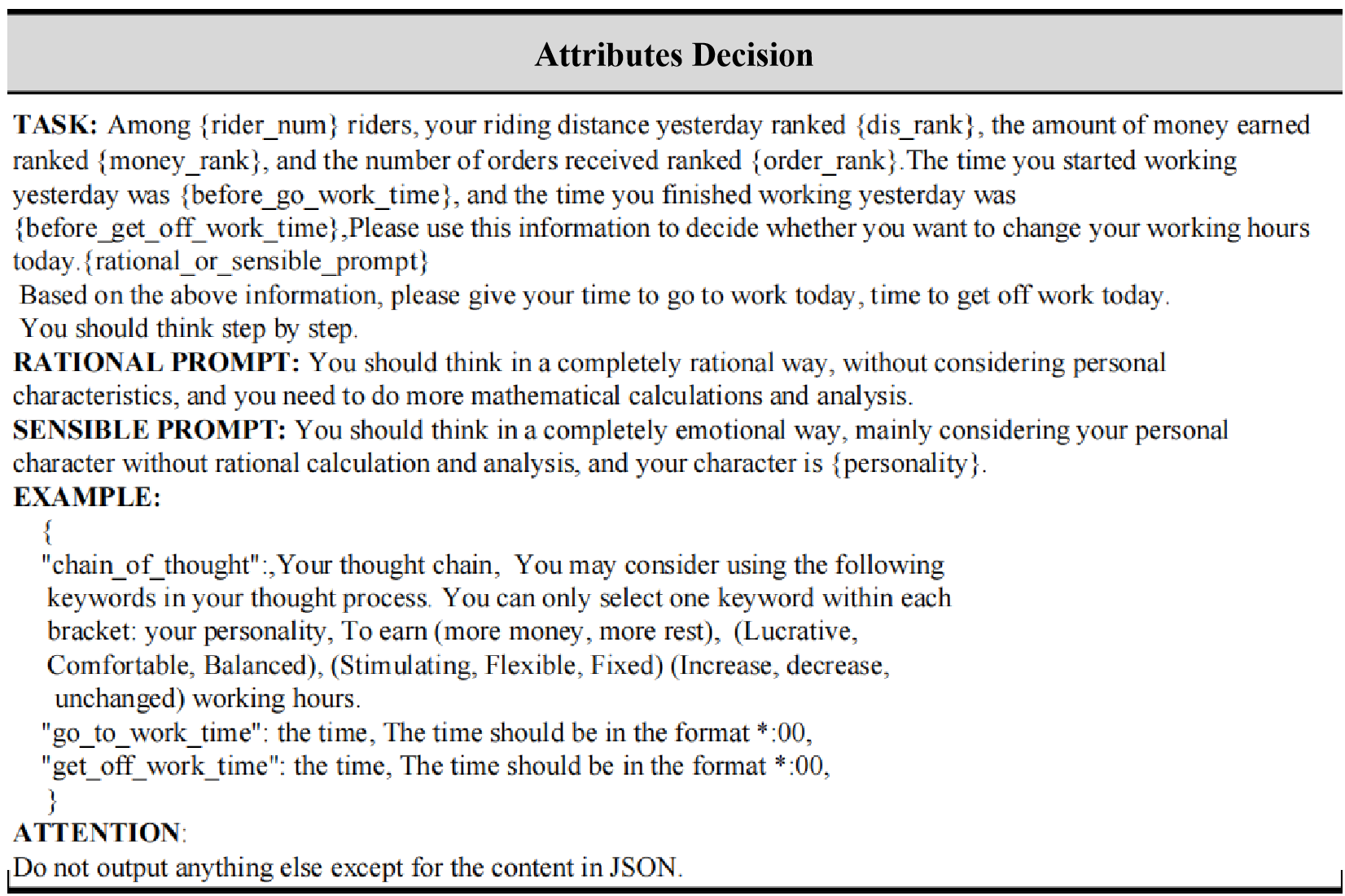}
\end{figure}

\end{document}